\documentclass[a4paper,conference]{IEEEtran}

\usepackage{amsmath,amssymb,amsfonts}
\usepackage{algorithmic}
\usepackage[backend=bibtex, sorting=none, style=ieee]{biblatex}
\usepackage{graphicx}
\usepackage{import}
\usepackage{textcomp}
\usepackage{xcolor}
\usepackage{array}
\usepackage{booktabs}
\ifCLASSOPTIONcompsoc
\else
\fi

\ifCLASSINFOpdf
\else
\fi
\usepackage{xspace}
\usepackage{svg}
\usepackage[ruled,vlined]{algorithm2e}
\usepackage[T1]{fontenc}
\usepackage{subcaption, booktabs}

\newcommand{\namealgo}{\textsc{PIF}\xspace} 

\newcommand{\ifor}{\textsc{iFor}\xspace}
\newcommand{\eifor}{\textsc{EiFor}\xspace}
\newcommand{\lof}{\textsc{LOF}\xspace}
\newcommand{\ptree}{PI-\textsc{Tree}\xspace}
\newcommand{\ptrees}{PI-\textsc{Trees}\xspace}
\newcommand{\pforest}{PI-\textsc{Forest}\xspace}

\DeclareRobustCommand{\vect}[1]{
  \ifcat#1\relax
    \boldsymbol{#1}
  \else
    \mathbf{#1}
  \fi}

\title{PIF: Anomaly detection via preference embedding}

\author{
    \IEEEauthorblockN{Filippo Leveni}
    \IEEEauthorblockA{Politecnico di Milano (DEIB)\\
                      filippo.leveni@polimi.it}
    \and
    \IEEEauthorblockN{Luca Magri}
    \IEEEauthorblockA{Politecnico di Milano (DEIB)\\
                      luca.magri@polimi.it}
    \and
    \IEEEauthorblockN{Giacomo Boracchi}
    \IEEEauthorblockA{Politecnico di Milano (DEIB)\\
                      giacomo.boracchi@polimi.it}
    \and
    \IEEEauthorblockN{Cesare Alippi}
    \IEEEauthorblockA{Politecnico di Milano (DEIB)\\
                      cesare.alippi@polimi.it\\
                      Università della Svizzera italiana\\
                      cesare.alippi@usi.ch}
}

\begin{document}
    \maketitle
    \begin{abstract}
        We address the problem of detecting anomalies with respect to structured patterns. 
To this end, we conceive a novel anomaly detection method called \namealgo, that combines the advantages of adaptive isolation methods with the flexibility of preference embedding. Specifically, we propose to embed the data in a high dimensional space where an efficient tree-based method, \pforest, is employed to compute an anomaly score.
Experiments on synthetic and real datasets demonstrate that \namealgo favorably compares with state-of-the-art anomaly detection techniques, and confirm that \pforest is better at measuring arbitrary distances and isolate points in the preference space.



    \end{abstract}
    
    \section{Introduction}
        \label{sec:introduction}
\emph{Anomaly detection} deals with the problem of identifying data that do not conform to an expected behavior \cite{ChandolaBanerjeeAl09}. This task, sometimes referred to as outlier detection, finds numerous applications in fraud \cite{AhmedMahmoodAl16}, \cite{DalPozzoloBoracchi18} 
and intrusion \cite{BronteShahriarAl16} detection, health \cite{BanaeeAhmedAl13} 
and quality monitoring \cite{StojanovicDinicAl16}
, to name a few examples.
In the statistical and data-mining literature, \emph{anomalies} are typically detected as samples falling in low-density regions of a probability density model describing the data \cite{ChandolaBanerjeeAl09}. On the contrary, \emph{normal} 
data are samples that lie in denser regions.
In this paper, we consider anomaly detection in a pattern-recognition 
setup, where anomalies are samples that deviate from certain structured patterns.
Although \emph{statistical anomalies} can also be seen as a particular case of these \emph{pattern-recognition anomalies}, where the model describing normal data is a pdf, statistical-based and pattern-based approaches are traditionally treated separately in the literature as they employ different algorithms and methods.

\begin{figure}
    \centering
    \subfloat[Density\label{subfig:density}]{\includegraphics[width = 0.25\linewidth]{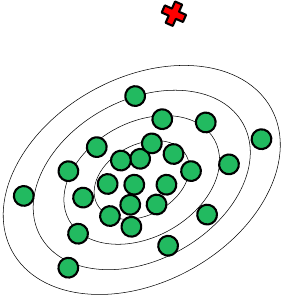}}
    \qquad
    \subfloat[Pattern\label{subfig:model}]{\includegraphics[width = 0.25\linewidth]{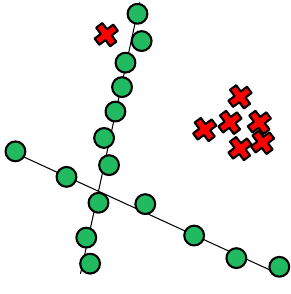}}
    \caption{Left: an anomaly (marked as \textcolor{red}{$\times$}) is recognized as a point in a low density area. Right: anomalies are defined with respect to their deviation from patterns described by line equations.
    }
    \label{fig:toyParametric}
\end{figure}

Fig.~\ref{fig:toyParametric} illustrates differences between statistical and pattern-recognition anomalies. In Fig.~\ref{subfig:density} a statistical anomaly can be easily identified as a sample falling in a low density area, in Fig.~ \ref{subfig:model} anomalies are instead points that are not collinear, while normal data belong to two patterns described by line equations. Note that density by itself in this latter case is not meaningful to identify anomalies, unless further processing is considered. 
Although overly simplified, Fig.~\ref{subfig:model} illustrates a primary task that has to be successfully addressed in several computer vision and pattern recognition applications, where more general parametric models are used instead of straight lines to identify structures or regularities in data.
Finding anomalies with respect to a parametric model is at the core of many low level vision tasks, like robust curve detection \cite{XuOjaAl90}. Moreover, this is a problem routinely addressed in Structure-from-Motion \cite{FarenzenaFusielloAl09}, where data consist in point-wise matching features between multiple images, and anomalies are wrong matches that cannot be described by consistent geometric transformations, such as homographies or fundamental matrices.
Other examples include 3D registration \cite{BeslMcKay92}, where anomalous matches are defined with respect to rototranslations, and object/template matching \cite{Brunelli09, SushkovSammut10, ChoLeeAl09}.
Anomaly detection in these settings is very challenging, since anomalies cannot be directly removed without having identified each and every structure first but, at the same time, anomalies hinder the identification of the existing structures.
For this reason, anomalies are  often detected as a byproduct of a multi-structure estimation process, which is performed through robust model fitting algorithms~\cite{FischlerBolles81, ToldoFusiello08, MagriFusiello14, MagriFusiello16}. Within this framework, the structures underlying normal data are first identified, and then, all those points that do not conform with them are labeled as anomalous.
We believe that model fitting is by far a more difficult problem than anomaly detection and that algorithms directly detecting anomalies would be preferable in those situations where anomaly detection is the primary goal (e.g., because anomalies convey relevant information on their own). Even in those situations where the focus is on the recovery of structures/models, it might be convenient to eliminate structure-less samples first, to ease the subsequent structure estimation task, as demonstrated by several domain-specific pre-filtering techniques  employed before robust estimation (e.g.,~\cite{CechMatasAl10,SattlerLeibeAl09}).

Here we present Preference Isolation Forest (\namealgo), a novel algorithm to directly detect anomalies among structures  whose nature is described by a given parametric function of unknown parameters.
To this purpose, we embed data into an high dimensional space, called \emph{preference space}~\cite{ToldoFusiello08,MagriFusiello14}, and then we perform anomaly detection by relying on \pforest, an efficient tree-based method that reflects a suitable distance metric for identifying anomalies.
To the best of our knowledge, this is the first time anomaly detection is applied in the preference space.
Extensive experiments show that (i) exploiting structure information allows to identify anomalies effectively and (ii)  \namealgo outperforms state-of-the-art anomaly detectors like Local Outlier Factor~\cite{BreunigKriegel00} (\lof), Isolation Forest~\cite{LiuTingAl08, LiuTingAl12} (\ifor) and Extended Isolation Forest~\cite{HaririKindAl18} (\eifor).
Most remarkably, we show that straightforward solutions plugging-in anomaly detection algorithms in the preference space are often not successful. We speculate that this is due to the fact that these methods do not leverage an appropriate distance function for the preference space. On the contrary, \pforest achieves superior performance thanks to a nested Voronoi tessellations constructed over the Tanimoto distance \cite{Tanimoto57} that is specifically designed to capture preference agreements.

In summary our main contributions are:
\begin{itemize}
    \item \namealgo, the first algorithm that identifies anomalies with respect to structured patterns by means of preference embedding.
    \item \pforest, a novel tree-based anomaly detection tool that is very successful in the preference space, but that can be extended to any metric space.
\end{itemize}
\par

    \section{Problem Formulation}
        \label{sec:problem_formulation}
        We assume we are given a noisy finite dataset $X = \{\vect{x}_1,\ldots, \vect{x}_n\}\subset \mathbb{R}^d$ 
containing both normal\footnote{Herein and through the paper by normal we do not mean Gaussian, but rather that conform to the normal state.} and anomalous data. Normal data can be described as belonging to a union of structures $S = S_1 \cup \ldots \cup S_k$, defined by a parametric model family $\mathcal{F}$, of which we assume to know the analytical expression. In particular, each structure $S_i$  corresponds to the vanishing set of an instance of $\mathcal{F}$, described by a specific parameter vector $\vect{\theta}_i$. In a noisy free setup, all normal points must satisfy equation $\mathcal{F}(\vect{x},\vect{\theta}_i) = 0$ for some parameter vector $\vect{\theta}_i$ but, because of noise, the previous equation is not necessarily satisfied, and we should rather expect $\mathcal{F}(\vect{x},\vect{\theta}_i) \approx 0$ for every $\vect{x} \in S_i$. Anomalies instead do not refer to any structure, and form a subset $A \subset X$.
For example, the inliers depicted in Fig.~\ref{subfig:model} can be described as a collection of two lines: in this case, each normal point  $\vect{x}=(x,y)\in S\subset \mathbb{R}^2$ approximately satisfies a first degree equation $\mathcal{F}(\vect{x},\vect{\theta}) = \theta_1 x+\theta_2y +\theta_3 =0$, where $\vect{\theta}=(\theta_1,\theta_2,\theta_3)$ represents the line coefficients. 
On the contrary, anomalies are generated by a different unknown process and are not coherent with lines.

We address the problem of automatically detecting all the anomalies $A \subset X$.
Specifically, our aim is to derive an \emph{anomaly score} $\alpha\colon X\to \mathbb{R}$ that ranks higher the anomalies, namely $\alpha(\vect{a})\gg\alpha(\vect{s})$ for all $\vect{a}\in A$ and $\vect{s}\in S$, so that it is possible to detect them by setting an appropriate threshold.
This task is particularly challenging since the overall number of structures $k$ as well as their parameters $\{\vect{\theta}_i\}_{i=1,\ldots,k}$ are unknown. The nature and amount of noise determining how much normal data depart from the analytical equations of their corresponding structure is unknown, but we assume it can be directly estimated (e.g., \cite{WangSuter04}).







    \section{Related Work}
        \label{sec:related_works}
    Several approaches have been proposed to identify anomalies, and for a comprehensive description refer to~\cite{ChandolaBanerjeeAl09, ChandolaBanerjeeAl10}.
    A possible taxonomy envisages three main categories: distance-based, density-based and model-based.
    In distance-based anomaly detection \cite{KnorrNgAl00} an instance is considered anomalous when its neighborhood does not contain a sufficient number of samples. Simplest methods of this category are based on the K-Nearest Neighbors \cite{RamaswamyRastogiAl00} approach: the anomaly score of a data sample, is simply the distance to its $k$-th nearest neighbor. Better results can be obtained when data-dependent distance measures are employed \cite{TingZhuAl16}.
    Density-based anomaly detection methods (e.g., \cite{BreunigKriegel00,KriegelKrogerAl09,PapadimitriouKitigawaAl03}) follow a similar idea, but density is used instead of distance. The key concept is that anomalous and normal instances differ in their local density. An important algorithm representative of this category is \lof~\cite{BreunigKriegel00}. The basic idea is that density around a normal instance is similar to the density around its $k$-neighbors, in contrast the density around an anomaly is significantly different from the local density of its $k$-neighbors. 
    
    In model-based anomaly detection it is assumed that normal data are generated from a model, thus, the more an instance deviates from the model, the higher its probability to be anomalous.
    Most existing model-based methods learn a model from the data, then identify anomalies as those data points that do not fit the model well. Notable examples of this approach are classification-based methods~\cite{AbeZadroznyAl06}, reconstruction-based methods~\cite{ChenSatheAl17}, \cite{CarreraRossi19}, \cite{CarreraManganini17}, and clustering-based methods~\cite{HeXuAl03}.
    Isolation Forest (\ifor)~\cite{LiuTingAl08, LiuTingAl12} instead directly isolates anomalies by assuming that
    they are ``few and different''~\cite{LiuTingAl08} compared to normal instances.
    \ifor builds a forest of randomized trees~\cite{GeurtsErnstAl06} from data and~\cite{LiuTingAl08, LiuTingAl12} show that, on average, anomalous points end up in leaves at shallower levels of tree height than normal data (herein and hereafter we refer to \emph{height} as a synonym for \emph{depth}).
    An effective extension of \ifor is Extended Isolation Forest (\eifor)~\cite{HaririKindAl18}.
    \par
    The above methods are usually applied in the \emph{ambient space}, namely the space where data are given, and are effective to identify statistical anomalies. However, when dealing with pattern-recognition anomalies, this is not the best solution as pattern or group-level information is not exploited.
    In our approach we leverage on the principle that multiple normal points belong to a structure, shifting the problem in a preference space where anomalies can be easily separated from structured data. In the next section we recall the main concepts of preference analysis necessary for this construction.

    \section{Preference Isolation Forest}
        \label{sec:iVor}
        
The proposed Preference Isolation Forest (\namealgo) computes anomaly scores $\alpha$ in two main steps:
(i) embedding the data in the \emph{preference space}  and  (ii)  adopting a tree-based isolation approach to detect anomalies in the preference space.

\begin{algorithm}[tb]
    \caption{\namealgo anomaly detection \label{alg:main}}
    \DontPrintSemicolon
    \SetNoFillComment
    \KwIn{$X$ - input data, $t$ - number of trees, $\psi$ - sub-sampling size, $b$ - branching factor}
    \KwOut{Anomaly scores $\{\alpha_\psi(\mathcal{E}(\vect{x}_i))\}_{i=1,\ldots,n}$}
    \begin{small}
        \tcc{Preference embedding}
    \end{small}
    Sample $m$ models $\{\vect{\theta}_i\}_{i=1,\ldots,m}$ from $X$ \label{line:begin_embedding}\\
    $P \leftarrow preferenceEmbedding(X, \{\vect{\theta}_i\}_{i=1,\ldots,m})$ \label{line:end_embedding}\\
    \begin{small}
        \tcc{Training Preference Isolation Forest}
    \end{small}
    $F \leftarrow \text{\pforest}(P, t, \psi, b)$ \label{line:begin_detection}\\
    \begin{small}
        \tcc{Scoring input data}
    \end{small}
    \For{$i = 1$ to $|P|$}
        {$\vect{h} \leftarrow [0, \ldots, 0] \in \mathbb{R}^t$\\
        \For{$j = 1$ to $t$}
             {$T \leftarrow \text{$j$-th \ptree in $F$}$ \\ 
              $[\vect{h}]_j \leftarrow \text{\textsc{PathLength}}(\vect{p}_i, T, 0)$}
        $\alpha_\psi(\vect{p}_i) \leftarrow 2^{-\frac{E(\vect{h(\vect{p}_i)})}{c(\psi)}}$}
    return $\{\alpha_\psi(\vect{p}_i)\}_{i=1,\ldots,n}$ \label{line:end_detection}\\
\end{algorithm}


\subsection{Preference embedding \label{sec:pref}}

For the first time in the context of anomaly detection, we propose to use the preference embedding, a technique previously used in the multi-model fitting literature \cite{ToldoFusiello08,MagriFusiello14,MagriFusiello16} to which the interested reader is referred for further details. \namealgo starts by mapping each point $\vect{x}\in X$ to an $m$-dimensional vector having components in the unitary interval $[0,1]$, via a mapping  $\mathcal{E}\colon X\to [0,1]^m$. The space $[0,1]^m$ is called preference space.
More precisely, the embedding depends on: a family $\mathcal{F}$ of models parametric in $\vect{\theta}$, a set of $m$ model instances $\{\vect{\theta}_i\}_{i=1,\ldots,m}$ and an estimate of the standard deviation $\sigma$ of the noise affecting the data. A sample $\vect{x}_i\in X$ is then embedded to a vector $\vect{p}_i = \mathcal{E}(\vect{x}_i)$ whose $j$-th component is defined as
\begin{equation}
\label{eq:preference}
[\vect{p}_i]_j  =  \begin{cases}
\phi(\delta_{ij}) &\text{if $\delta_{ij}=\mathcal{F}(\vect{x}_i,\vect{\theta}_{j})\leq 3\sigma$ }\\
0 &\text{otherwise}
\end{cases},
\end{equation}
where $\delta_{ij}=\mathcal{F}(\vect{x}_i,\vect{\theta}_j)$ measures the deviation  of sample $\vect{x}_i$ with respect to the model $\vect{\theta}_{j}$, and $\phi$ is a monotonically decreasing function in $[0,1]$  such that $\phi(0)=1$. As in \cite{MagriFusiello16}, we use a Gaussian function of the form
\begin{equation}
\phi(\delta) = \exp(-\delta^2/\sigma).
\end{equation}
The $j$-th component of the preference vector $\vect{p}_i$, namely $[\vect{p}_i]_j$, is the preference granted by  a point $\vect{x}_i$ to model $\vect{\theta}_j$: the closer $\vect{x}_i$ to $\vect{\theta}_j$, the higher the preference.
The embedding function $\mathcal{E}$ maps dataset $X$ to the set of preference vectors
\begin{equation}
P =\{\vect{p}_i = \mathcal{E}(\vect{x}_i) \,|\, \vect{x}_i \in X \},
\end{equation}
which represents the image of $X$ through the embedding $\mathcal{E}$. 
The pool $\{\vect{\theta}_i\}_{i=1,\ldots,m}$ of $m$ models are sampled from the data using a RanSaC-like strategy (line 1, Algorithm \ref{alg:main}):
minimal sample set -- composed by the minimal number of points necessary to constraint a parametric model -- are extracted uniformly from the data, and are   used to determine the model parameters. For example, two points are drawn to determine the equation of a line $\vect{\theta}_j$.

The preference space is equipped with the Tanimoto distance \cite{Tanimoto57} to measure similarity between preferences: given two samples $\vect{p}_i = \mathcal{E}(\vect{x}_i)$ and $\vect{p}_j = \mathcal{E}(\vect{x}_j)$ their Tanimoto distance is 
\begin{equation}
\label{eq:tani}
\tau(\vect{p}_i,\vect{p}_j) = 1- \frac{\langle \vect{p}_i,\vect{p}_j\rangle}{\|\vect{p}_i\|^{2}+\|\vect{p}_j\|^{2}- \langle \vect{p}_i,\vect{p}_j\rangle}.
\end{equation}
Our choice for the Tanimoto distance is motivated by the fact that points conforming to the same structures share similar preferences, yielding low distances. In contrast, anomalies would result in null preferences to the majority of structures, thus resulting in very sparse vectors that tend to have distance close to $1$ with the majority of other samples.

\subsection{\pforest \label{sec:pforest}}
By moving from a Euclidean space to the preference space, state-of-the-art data driven methods such as \ifor and its variants perform poorly (as we will show in our experiments), since the splitting criterion used by these techniques implicitly measures point distances by the $\ell_2$ norm. For this reason, we build our solution upon \ptree, a novel tree-based isolation technique where the splitting criterion is based on Voronoi tessellations.
A Voronoi tessellation is simply a partition of a metric space into $b$ regions defined by $b$ samples, called seeds $\mathcal{S}=\{\vect{s}_i\}_{i=1,\ldots,b}$. The $i$-th region produced by the tessellation contains all the points $\vect{p}$ of the space having $\vect{s}_i$ as the closest seed in $\mathcal{S}$. The proposed \ptree is a nested version of Voronoi tessellations, where each region is further split in $b$ sub-regions; this procedure can be repeated recursively as illustrated in Fig.~\ref{fig:pifTree}. 
Voronoi tessellation as partitioning criterion preserves, better than other splitting schemes, the notion of distance in the preference space. In fact, in our case, regions are defined by seeds $\mathcal{S}$ and Tanimoto distance (Eq. \ref{eq:tani}). In general, Voronoi tessellation naturally applies to any metric space, preference space included.
    \begin{figure}
        \centering
        \includegraphics[width = 0.8\linewidth]{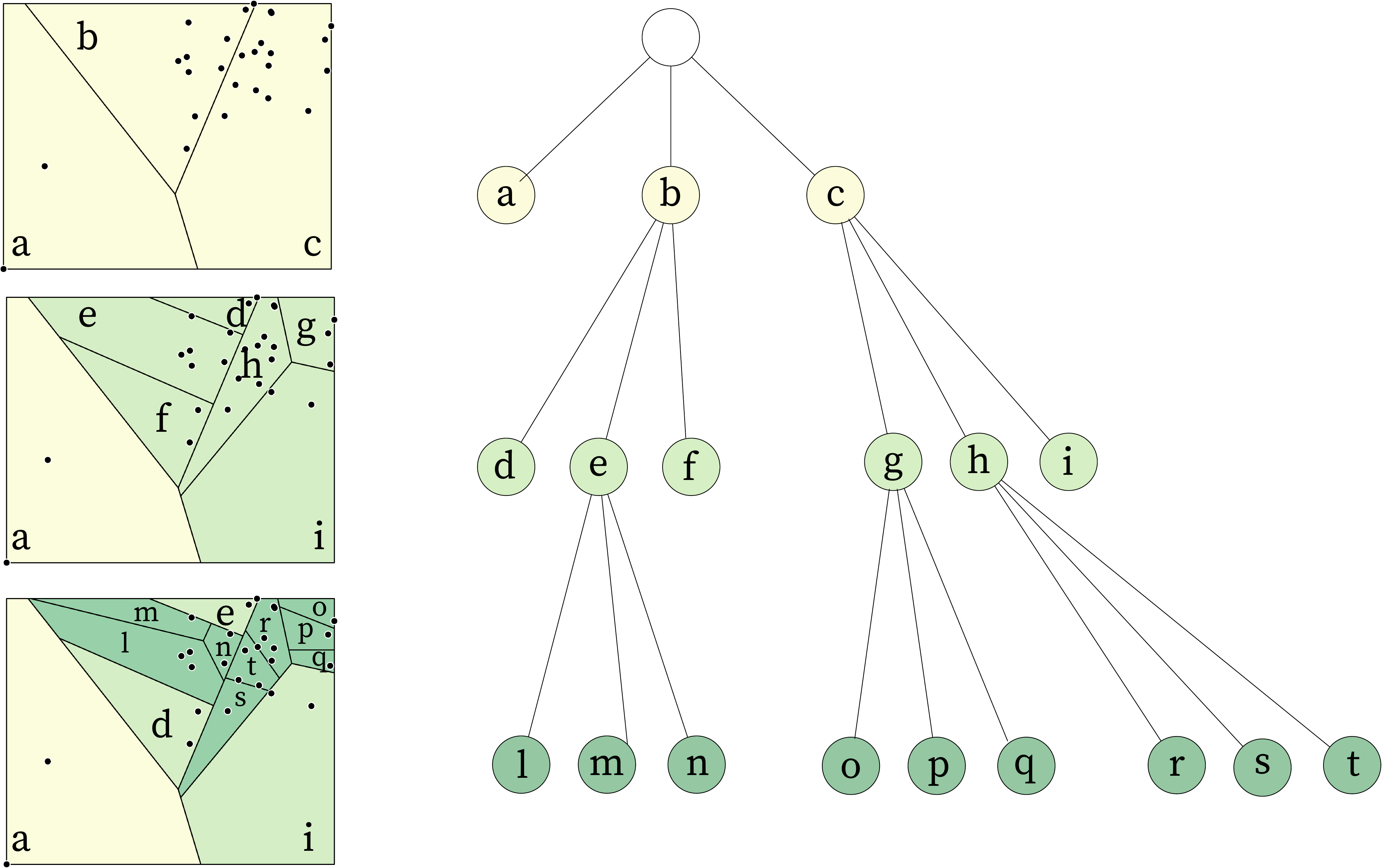}
        \caption{A \ptree with branching factor $b=3$ and height limit $l=3$ constructed from a set of points in $\mathbb{R}^2$. Every region is recursively split in $b$ sub-regions. The most isolated samples fall in leaves at lowest heights, such as 'a' and 'd' cells.}
        \label{fig:pifTree}
    \end{figure}
The construction of a \ptree is described in Algorithm~\ref{alg:iVorTree} and starts from a single region corresponding to the whole space $[0,1]^m$ that is then recursively split in $b$ sub-regions by randomly selecting $b$ seeds $\{\vect{s}_i\}_{i=1,\ldots,b} \subset P$ (line 4). Thus the points are partitioned into $b$ subsets $\mathcal{P}= \{P_i\}_{i=1,\ldots, b}$, each $P_i \subset P$ collecting those points in $P$ that have $\vect{s}_i$ as the closest seed according to Tanimoto distance (line 5). The number of seeds $b$ is the branching factor of the tree associated to the splitting process. The partitioning process stops when it is not possible to further split a region (i.e., the number of points in the region is less than $b$), or the tree reaches a maximum height (lines 1-2), set by default at $l = \log_{b}\psi$ (an approximation for the average tree height~\cite{Knuth98}), where $\psi$ is the number of points used to build the tree. The recursive process is outlined in lines \ref{line:begin_recursive}-\ref{line:end_recursive}, where a sub-tree is build for each subset $P_i \in \mathcal{P}$ (line 8).
    
    
     \begin{algorithm}[tb]
        \caption{\ptree \label{alg:iVorTree}}
        \KwIn{$P$ - preference representations, $e$ - current tree height, $l$ - height limit, $b$ - branching factor}
        \KwOut{A \ptree}
            \eIf{$e \geq l$ or $|P| < b$}
                {return $exNode\{Size \leftarrow |P|\}$}
                {randomly select a set of $b$ seeds $\{\vect{s}_i\}_{i=1,\ldots,b} \subset P$\\
                 $\mathcal{P} \leftarrow voronoiPartition(P, \{\vect{s}_i\}_{i=1,\ldots,b})$\label{line:split}\\
                 $chNodes \leftarrow \emptyset$ \label{line:begin_recursive}\\
                 \For{$i = 1$ to $b$}
                     {$chNodes \leftarrow chNodes \cup \text{\ptree}(P_i, e + 1,$\\
                      \hspace{4.9cm} $l, b)$} \label{line:end_recursive}
                return $inNode\{ChildNodes \leftarrow chNodes,$\\
                \hspace{2.2cm} $SplitPoints \leftarrow \{\vect{s}_i\}_{i=1,\ldots,b}\}$}
    \end{algorithm}

    \begin{algorithm}[tb]
        \caption{\pforest \label{alg:iVorForest}}
        \KwIn{$P$ - preference representations, $t$ - number of trees, $\psi$ - sub-sampling size, $b$ - branching factor}
        \KwOut{A set of $t$ \ptrees}
            $F \leftarrow \emptyset$\\
            set height limit $l = \log_{b}\psi$\\
            \For{$i = 1$ to $t$}
                {$P' \leftarrow subSample(P, \psi)$ \label{line:subsample}\\
                 $F \leftarrow F \cup \text{\ptree}(P', 0, l, b)$} \label{line:ensemble}
            return $F$
    \end{algorithm}
    
    \begin{algorithm}[tb]
        \caption{\textsc{PathLength}  \label{alg:pathLength}}
        \KwIn{$\vect{p}$ - a sample, $T$ - a \ptree, $e$ - current path length}
        \KwOut{Path length of $\vect{p}$}
            \If{$T$ is an external node}
              {return $e + c(T.size)$}
            $childNode \leftarrow voronoiLocate(\vect{p}, T.splitPoints,$ \label{line:locate}\\
            \hspace{4.4cm} $T.childNodes)$\\
            return $PathLength(\vect{p}, childNode, e + 1)$ \label{line:recur}
    \end{algorithm}
    The height of the leaf in which a point falls into is directly related to its separability: a lower height corresponds to points that can be separated with few splits from the rest of the data.
    To get an intuition of the concept of separability consider Fig.~\ref{fig:pifTree} where, for visualization purposes, the Euclidean distance is being considered. Here anomalies correspond to samples that fall in leaves with lower height.  Conversely, samples in high-density regions fall in leaves with higher height since in denser regions the number of possible recursive splits is higher.
    In order to gain robustness and to decrease the variance due to randomness in \ptree realizations, this idea is extended to \pforest, a forest  of \ptrees, and the \emph{average} height of a point in this forest is used to compute its overall anomaly score $\alpha$. 
    Algorithm~\ref{alg:iVorForest} details the construction of a \pforest containing
    $t$ \ptrees. Each \ptree is instantiated on a subset $P' \subset P$ of preference representations of $X$ (line~\ref{line:subsample}).
    The subsampling factor is controlled by the parameter $\psi$.

        
    \subsection{Anomaly score}
        \label{subsec:test}
        Anomaly scores are computed as in \ifor and other tree-based isolation methods~\cite{LiuTingAl08, LiuTingAl12, MensiBicego19}. With reference to Algorithm~\ref{alg:main}, after the samples are embedded in the preference space (lines 1-2) and the \pforest is built (line 3), each instance $\vect{p}\in P$ is passed through all the \ptrees of the \pforest, and the heights reached in every tree are computed and collected in a vector $\vect{h}(\vect{p}) = [h_{1}(\vect{p}), \ldots, h_{t}(\vect{p})]$ (lines 5-9).
        Then, (line 10) the anomaly score $\alpha$ is 
        \begin{equation}
            \alpha_\psi(\vect{p}) = 2^{-\frac{E(\vect{h}(\vect{p}))}{c(\psi)}},
        \end{equation}
        where $E(\vect{h}(\vect{p}))$ is the mean value over the elements of $\vect{h}(\vect{p})$ and $c(\psi)$ is an adjustment factor.
        
         The heights are computed through \textsc{PathLength} function described in Algorithm \ref{alg:pathLength}.
         In particular, at line \ref{line:locate}, the instance $\vect{p}$ is located in the corresponding region of the Voronoi partition having $T.splitPoints$ as seeds. Then the child node associated to the region is identified and used for the subsequent recursive process at line \ref{line:recur}.
         At line 2, the height is computed as $h_{i}(\vect{p}) = e + c(T.size)$, that is the height $e$ of the leaf where $\vect{p}$ falls in the $i$-th tree, plus an adjustment coefficient $c(n)$ that depends on the cardinality $n$ of this leaf. The adjustment factor is necessary to take into account subtrees that stopped the construction process, having reached the height limit $l$. We assume that $b=2$, in this case the adjustment factor becomes as \cite{LiuTingAl08, LiuTingAl12}:
        \begin{equation}
            c(n) =
            \begin{cases}
                0 &\text{if $n = 1$}\\
                1 &\text{if $n = 2$}\\
                2H(n - 1) - 2(n - 1)/n &\text{if $n > 2$}
            \end{cases},
        \end{equation}
        where $H(i)$ is the harmonic number and it can be estimated by $\textrm{ln}(i) + \gamma$ (being $\gamma$ the Euler's constant).

        The complexity of \namealgo is $O( \psi \cdot t \cdot b\cdot \log_{b}\psi)$ as it regards \pforest construction, and $O(n\cdot t\cdot b\cdot\log_{b}\psi)$ for the scoring phase, where $n$ is the number of instances to be scored. With respect to \ifor we have an additional overhead due to the embedding $\mathcal{E}(\cdot)$, which however can be easily parallelized.

    \label{subsec:remark}
    
    \namealgo has two main advantages over other isolation-based anomaly detection tools, like \ifor. First, the  preference trick allows to integrate useful information about normal data and consequently to better characterize structure-less anomalies.
    Second, the Voronoi tessellation preserves the intrinsic distance of the preference space (i.e., Tanimoto distance) during the splitting process.
    These two features can be appreciated in the example reported in Fig. \ref{fig:poc} where the problem of identifying anomalies with respect to two circles is addressed. By looking at the color-coded anomaly scores produced by \ifor, \eifor, \lof, and \namealgo, it is possible to recognize that only \namealgo correctly identifies anomalies inside the circles.
    On the contrary \ifor, that works directly with $X$, does not reflect the structures of normal data, as it performs splits parallel to the axes, and consequently struggles to adapt to geometries that are not aligned to the main axes. With \eifor the situation improves slightly, but anomalies still cannot be precisely identified. Also \lof, which heavily depends on the neighbourhood parameter $k$ to estimate local density, has difficulties. \lof identifies as anomalous those regions where data density changes. The region outside the circles, although sparser, does not give rise to a change in density and it is thus erroneously identified as normal. Note that only \namealgo exploits the preference space, the other methods are confined to operate in terms of density and isolability. However, in the experiments reported in the next section, we will see that the advantages of \namealgo are not just motivated by the preference embedding, since other anomaly detection methods plugged in the preference space would yield lower detection performance than \namealgo. This further demonstrates the importance of \pforest having regions defined by Tanimoto distance rather than Euclidean distance.

    \begin{figure}[tb]
        \captionsetup[subfigure]{justification=centering}
        \centering
        \subcaptionbox{\vspace*{-4.6em}Input data}{
            \vspace*{-3.2em}
            \label{fig:input}
            \includegraphics[width=0.3\linewidth]{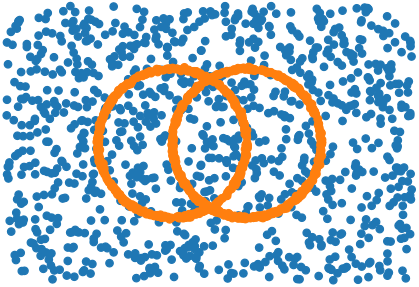}}
        \subfloat[\ifor
        \label{fig:ifor}]{
        \includegraphics[width=0.3\linewidth]{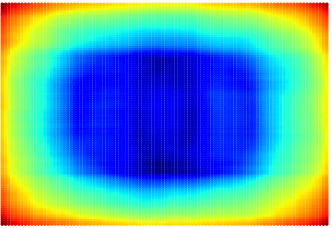}
        }
        \subfloat[\eifor
        \label{fig:eifor}]{
        \includegraphics[width=0.3\linewidth]{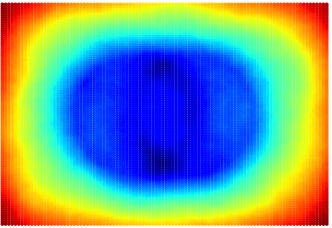}
        }
        \\
        \hfill
        \subfloat[\lof
        \label{fig:lof}]{
        \includegraphics[width=0.3\linewidth]{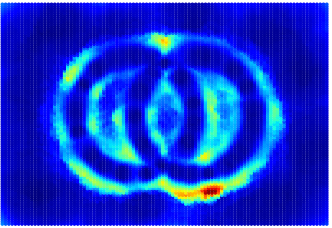}
        }
        \subfloat[\namealgo
        \label{fig:ivor_tanimoto}]{
        \includegraphics[width=0.3\linewidth]{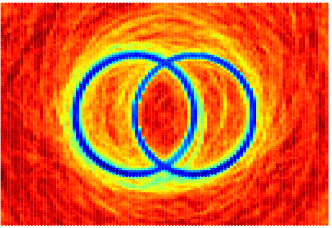}
        }
        \caption{Color-coded anomaly scores produced by different algorithms: high scores in red, low scores in blue. \label{fig:poc}}
        \label{fig:illustrative}
    \end{figure}

    \section{Experimental Validation}
        \label{sec:experimental_validation}
        We validate the advantages of \namealgo by assessing the benefits of coupling the preference trick with \pforest, both in terms of detection accuracy and stability. To this purpose, we adopt both synthetic and real-world datasets.

\subsection{Datasets}
    We consider the synthetic 2D datasets depicted in Fig. \ref{fig:synth2d}. The structures $S$ characterizing normal data are lines in stair$[*]$ and star$[*]$, and circles in circle$[*]$, where $[*]$ indicates the number of structures. Anomalies have been sampled from a uniform distribution within the bounding box containing normal data. For all the datasets every structure has 50 normal points, with the exception of stair3 and circle3 that have unbalanced structures, as detailed in Table~\ref{tab:synth2d}.
   Experiments on real data  are performed on the AdelaideRMF dataset~\cite{WongChinAl11}, that consists in stereo images and annotated matching points. Erroneous matches are also annotated and correspond to anomalies. The first 19 sequences refer to static scenes containing several planes, each giving rise to a set of matches described by an homography. For the remaining 19 sequences the scene is not static: several objects move independently and give rise to a set of point correspondences described by a a fundamental matrix. In both these scenarios we want to recognize anomalous matches.
   
       \begin{figure*}[htb]
        \captionsetup[subfigure]{justification=centering}
        \centering
        \subfloat[stair3]{
        \includegraphics[height=0.11\linewidth]{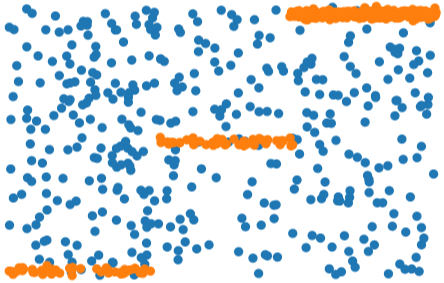}
        }
        \subfloat[stair4]{
        \includegraphics[height=0.11\linewidth]{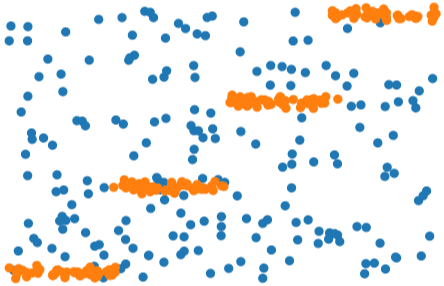}
        }
        \subfloat[star5]{
        \includegraphics[height=0.11\linewidth]{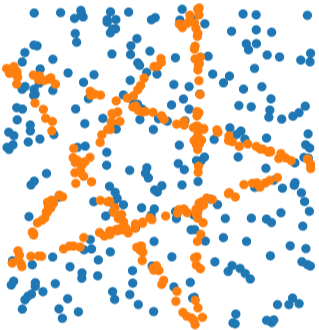}
        }
        \subfloat[star11]{
        \includegraphics[height=0.11\linewidth]{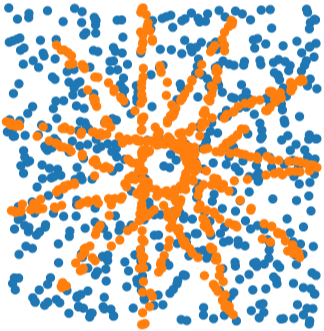}
        }
        \subfloat[circle3]{
        \includegraphics[height=0.11\linewidth]{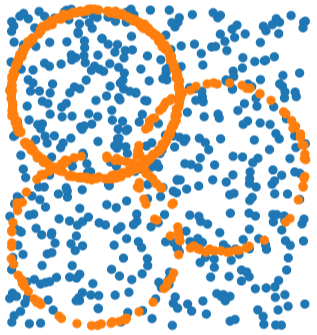}
        }
        \subfloat[circle4]{
        \includegraphics[height=0.11\linewidth]{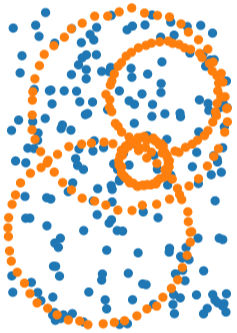}
        }
        \subfloat[circle5]{
        \includegraphics[height=0.11\linewidth]{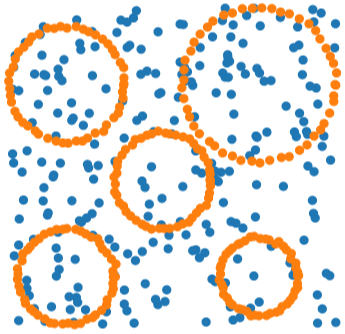}
        }
        \caption{Synthetic datasets. Orange dots represent normal data, while blue dots represent anomalies.}
        \label{fig:synth2d}
    \end{figure*}

\subsection{Methodology}
    
    We compare \namealgo with \ifor, \eifor and \lof. To assess the benefits of the preference trick, experiments on synthetic data are performed in the (i) ambient (Euclidean) space, (ii) preference space and (iii) binarized preference space, where the Tanimoto distance specializes exactly to the Jaccard distance \cite{Jaccard01} (binary vectors are used in Eq. \eqref{eq:tani}).
    We also explore the performance of \namealgo without the preference embedding (\namealgo $\ell_2$), i.e. setting $P = X$, and with the binarized preference embedding (\namealgo jac). 
    Preferences are computed with respect to a pool of $m=10|X|$ model instances, being circles used in circle$[*]$ datasets and lines elsewhere.
    
    To evaluate the stability of our approach, we perform an additional experiment on stair3 and circle5. These datasets have been modified so that they contain different percentages of anomalies. In particular, for both datasets, $|X|$ is kept fixed and equal to $1000$, while $|S|$ and $|A|$ vary as: $\frac{|A|}{|X|} = \{0.05, 0.1, 0.2, 0.3, 0.4, 0.5, 0.6, 0.7, 0.8 , 0.9\}$. 
    
    Experiments on real datasets are performed in the preference space. The pool of models is determined by sampling $m = 6|X|$ model instances, homographies for the first 19 datasets and fundamental matrices for the remaining 19.

    The parameters of \ifor, \eifor and \namealgo are kept fixed to $t= 100, \psi = 256$ and $b = 2$ in all the experiments.
    As regard \lof instead, given its sensitivity to the neighborhood size, various values of the parameter $k$ are employed, ranging from $k = 10$ to $k = 500$.
 
    Anomaly detection performance, collected in Tab.~\ref{tab:lines_and_circles},~\ref{tab:homographies} and~\ref{tab:fundamentals}, are evaluated in terms of AUC averaged over 10 runs.
    The highest value for each dataset is underlined, whereas a boldface indicates that the best AUC is statistically better than its competitors in the same embedding, according to a paired t-test with $\alpha = 0.05$. 
    The \lof results refer to the $k$ that maximizes the average AUC along the datasets of each model family, for all the embeddings ($k$ parameter values are reported in Table~\ref{tab:neighborhood}).
    Fig. \ref{fig:propOutliers} displays the AUC achieved in correspondence with different percentages of anomalies employed, averaged over 10 runs. Only the best 3 values of $k$ for \lof have been reported.
\subsection{Discussion}
    Table~\ref{tab:lines_and_circles} shows that all methods improve their performance when performed in the preference space. Therefore, it is always convenient to exploit the preference trick  rather than working directly in the ambient space (i.e., in $X$).  In addition, continuous preferences must be preferred over binary ones, as their greater expressiveness produces a space where anomalies can be identified more effectively. Note that \lof achieves a better AUC on star5, star11 and circle5 where normal data are evenly distributed among structures, and it is possible to get an optimal parameter of $k$. On the contrary, \lof performs poorly when structures have different cardinalities (stair3 and circle3). \namealgo instead works with fixed parameters and attains good results also for unbalanced structures; on average \namealgo is the best statistical significantly method.
    The critical dependence  of \lof to $k$ is also highlighted in Fig. \ref{fig:propOutliers} where stability is evaluated with respect to the rate of anomalies contamination.  As the percentage of anomalies varies, the optimal value of $k$ changes and it becomes difficult to guess a correct size for the neighborhoods of normal points. This is particularly evident in the unbalanced dataset stair3. As a consequence, the performance of \lof has a great variability according to the chosen $k$.
    Moreover, although \ifor and \eifor achieve more stable performances, they are not as stable as \namealgo. In addition, there is a very apparent performance gap between \ifor, \eifor and \namealgo, since only the latter is able to deal with the Tanimoto metric.
  
    As real data are concerned, even if the high AUC values shown in Tab.~\ref{tab:homographies} and~\ref{tab:fundamentals} suggest that anomaly detection on these tasks is easier than in the synthetic case, 
    the difference in performance between \ifor, \eifor (Euclidean distance) and \lof, \namealgo (Tanimoto distance) remains evident.
    \par
    Both \lof and \namealgo seem to be valid methods to perform anomaly detection in the preference space, although on average the performance of \namealgo is better: the difference between mean AUCs is always statistically in favor of \namealgo, except on Tab. \ref{tab:fundamentals} where the methods are statistically equivalent.
    In addition, \lof  
    must be tuned according to the expected size of the neighborhoods of normal points
   while \namealgo has been tested with fixed parameters.
 
    \begin{table}[]
        \begin{subtable}{.48\textwidth}
            \centering
            \resizebox{.8\textwidth}{!}{
            \begin{tabular}{lllll}
                \toprule
                        & $|X|$  & $|A|$ & $|S|$ & $|S_1|, \ldots, |S_k|$             \\
                \midrule
                stair3  & 800  & 400 & 400 & $|S_1| = 272$, $|S_i| = 64 \quad\forall i \in \{2, 3\}$ \\
                stair4  & 400  & 200 & 200 & $|S_i| = 50 \quad\forall i \in \{1, \ldots, 4\}$        \\
                star5   & 500  & 250 & 250 & $|S_i| = 50 \quad\forall i \in \{1, \ldots, 5\}$        \\
                star11  & 1100 & 550 & 550 & $|S_i| = 50 \quad\forall i \in \{1, \ldots, 11\}$       \\
                circle3 & 1000 & 500 & 500 & $|S_1| = 376$, $|S_i| = 62 \quad\forall i \in \{2, 3\}$ \\
                circle4 & 400  & 200 & 200 & $|S_i| = 50 \quad\forall i \in \{1, \ldots, 4\}$        \\
                circle5 & 500  & 250 & 250 & $|S_i| = 50 \quad\forall i \in \{1, \ldots, 5\}$        \\
                \bottomrule
            \end{tabular}
            }
            \caption{Synthetic datasets settings}
            \label{tab:synth2d}
        \end{subtable}
      
        \begin{subtable}{.48\textwidth}
            \centering
            \resizebox{.8\textwidth}{!}{
            \begin{tabular}{llll}
            \toprule
                    & Euclidean & Preference binary & Preference             \\
            \midrule
            circle  & 75  & 25 & 25 \\
            line  & 25 & 150 & 75 \\
            homography  & - & 100 & 100 \\
            fundamental  & - & 75 & 80 \\
            \bottomrule
            \end{tabular}
            }
            \caption{Optimal $k$ parameters of \lof}
            \label{tab:neighborhood}
        \end{subtable}
        \caption{Experiment settings}
    \end{table}
    
    \begin{table*}[]
        \centering
        \resizebox{.8\textwidth}{!}{
        \begin{tabular}{lllllllllllll}
            \toprule
                    & \multicolumn{4}{l}{Euclidean} & \multicolumn{4}{l}{Preference binary} & \multicolumn{4}{l}{Preference}  \\
                    \midrule
                    & \lof $\ell_2$  & \ifor  & \eifor & \namealgo $\ell_2$ & \lof jac  & \ifor  & \eifor & \namealgo jac & \lof tani & \ifor  & \eifor & \namealgo \\
                    \midrule
            stair3  & 0.737 & 0.925 & 0.920 & 0.918 & 0.904    & 0.885 & 0.864 & 0.958    & 0.815    & 0.923 & 0.925 & \textbf{\underline{0.971}}     \\
            stair4  & 0.814 & 0.889 & 0.874 & 0.871 & 0.849    & 0.855 & 0.860 & 0.941    & 0.881    & 0.912 & 0.908 & \textbf{\underline{0.952}}     \\
            star5   & 0.771 & 0.722 & 0.738 & 0.788 & 0.875    & 0.745 & 0.769 & 0.872    & \textbf{\underline{0.929}}    & 0.761 & 0.822 & 0.910     \\
            star11  & 0.671 & 0.728 & 0.727 & 0.738 & 0.830    & 0.739 & 0.741 & 0.771    & \textbf{\underline{0.900}}    & 0.738 & 0.774 & 0.796     \\
            circle3 & 0.761 & 0.698 & 0.732 & 0.779 & 0.719    & 0.842 & 0.854 & 0.900    & 0.731    & 0.854 & 0.891 & \textbf{\underline{0.930}}     \\
            circle4 & 0.640 & 0.641 & 0.665 & 0.679 & 0.827    & 0.686 & 0.699 & 0.860    & \underline{0.906}    & 0.667 & 0.720 & 0.897     \\
            circle5 & 0.543 & 0.569 & 0.570 & 0.633 & 0.699    & 0.597 & 0.617 & 0.672    & \textbf{\underline{0.823}}    & 0.573 & 0.593 & 0.780     \\
            \midrule
            Mean    & 0.705 & 0.739 & 0.747 & 0.772 & 0.815    & 0.764 & 0.772 & 0.853    & 0.855    & 0.775 & 0.805 & \textbf{\underline{0.891}}     \\
        \bottomrule
        \end{tabular}
        }
        \caption{Synthetic datasets AUCs}
        \label{tab:lines_and_circles}
    \end{table*}

    \begin{figure}
        \hspace{1.21em}
        \centering
        \includegraphics[width=0.89\linewidth]{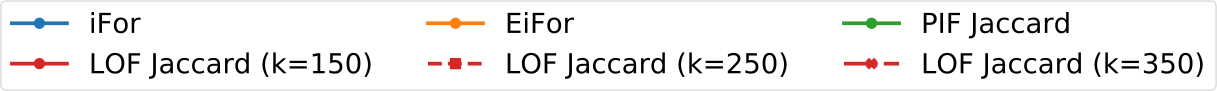}
        \\
        \begin{subfigure}{0.47\linewidth}
            \centering
            \includegraphics[width=\linewidth]{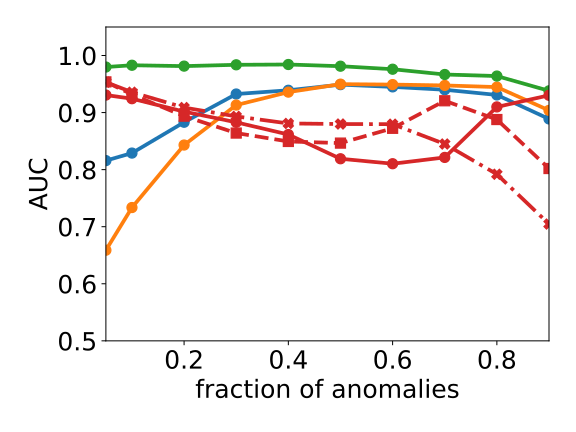}
            \caption{stair3, unbalanced structures}
            \label{fig:stair3}
        \end{subfigure}
        \hfill
        \begin{subfigure}{0.47\linewidth}
            \centering
            \includegraphics[width=\linewidth]{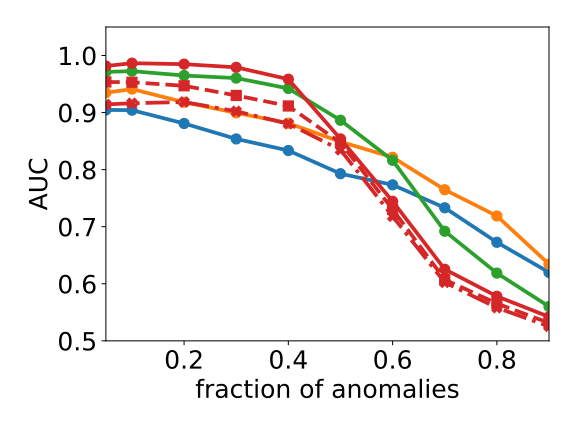}
            \caption{circle5, balanced structures}
        \end{subfigure}
        \caption{AUCs achieved at various percentages of anomalies.}
        \label{fig:propOutliers}
    \end{figure}
    \begin{table*}[]
        \begin{subtable}{.47\textwidth}
            \centering
            \resizebox{.795\textwidth}{!}{
            \begin{tabular}{lllll}
                \toprule
                                & \lof tani & \ifor  & \eifor & \namealgo \\
                                \midrule
                barrsmith       & \textbf{\underline{0.969}}    & 0.708 & 0.715 & 0.944     \\
                bonhall         & 0.918    & \underline{0.969} & 0.967 & 0.949     \\
                bonython        & \textbf{\underline{0.978}}    & 0.679 & 0.691 & 0.954     \\
                elderhalla      & \underline{0.999}    & 0.925 & 0.909 & \textbf{\underline{0.999}}     \\
                elderhallb      & 0.986    & 0.924 & 0.943 & \textbf{\underline{0.999}}     \\
                hartley         & 0.963    & 0.749 & 0.793 & \textbf{\underline{0.989}}     \\
                johnsona        & 0.993    & 0.993 & 0.993 & \textbf{\underline{0.998}}     \\
                johnsonb        & 0.776    & \underline{0.999} & 0.998 & \underline{0.999}     \\
                ladysymon       & 0.847    & 0.944 & 0.943 & \textbf{\underline{0.997}}     \\
                library         & \textbf{\underline{1.000}}    & 0.764 & 0.771 & 0.998     \\
                napiera         & 0.975    & 0.869 & 0.879 & \textbf{\underline{0.983}}     \\
                napierb         & 0.888    & 0.931 & 0.936 & \textbf{\underline{0.953}}     \\
                neem            & 0.985    & 0.896 & 0.906 & \textbf{\underline{0.996}}     \\
                nese            & \textbf{\underline{0.996}}    & 0.888 & 0.892 & 0.980     \\
                oldclassicswing & 0.936    & 0.923 & 0.943 & \textbf{\underline{0.987}}     \\
                physics         & 0.670    & 0.858 & 0.787 & \textbf{\underline{1.000}}     \\
                sene            & \textbf{\underline{0.997}}    & 0.698 & 0.731 & 0.988     \\
                unihouse        & 0.785    & 0.998 & 0.998 & \textbf{\underline{0.999}}     \\
                unionhouse      & \textbf{\underline{0.987}}    & 0.639 & 0.664 & 0.968     \\
                \midrule
                Mean            & 0.929    & 0.861 & 0.866 & \textbf{\underline{0.983}}     \\
                \bottomrule
            \end{tabular}
            }
            \caption{Homographies}
            \label{tab:homographies}
        \end{subtable}
        \begin{subtable}{.47\textwidth}
            \centering
            \resizebox{.83\textwidth}{!}{
            \begin{tabular}{lllll}
                \toprule
                                  & \lof tani & \ifor  & \eifor & \namealgo \\
                                  \midrule
                biscuit           & 0.976    & 0.994 & 0.996 & \textbf{\underline{1.000}}     \\
                biscuitbook       & \underline{1.000}    & 0.987 & 0.988 & \underline{1.000}     \\
                biscuitbookbox    & \textbf{\underline{1.000}}    & 0.990 & 0.989 & 0.996     \\
                boardgame         & \textbf{\underline{0.962}}    & 0.400 & 0.304 & 0.949     \\
                book              & 0.996    & \underline{1.000} & \underline{1.000} & \underline{1.000}     \\
                breadcartoychips  & \textbf{\underline{0.989}}    & 0.978 & 0.971 & 0.976     \\
                breadcube         & \textbf{\underline{1.000}}    & 0.998 & 0.998 & 0.999     \\
                breadcubechips    & \textbf{\underline{0.999}}    & 0.985 & 0.985 & 0.998     \\
                breadtoy          & 0.984    & \underline{0.999} & 0.998 & \underline{0.999}     \\
                breadtoycar       & \textbf{\underline{0.998}}    & 0.933 & 0.883 & 0.991     \\
                carchipscube      & \textbf{\underline{0.993}}    & 0.981 & 0.966 & 0.987     \\
                cube              & \underline{0.999}    & 0.970 & 0.982 & \underline{0.999}     \\
                cubebreadtoychips & \underline{0.990}    & 0.962 & 0.958 & 0.989     \\
                cubechips         & \textbf{\underline{1.000}}    & 0.995 & 0.994 & \underline{1.000}     \\
                cubetoy           & \textbf{\underline{1.000}}    & 0.997 & 0.995 & \underline{1.000}     \\
                dinobooks         & 0.887    & 0.873 & 0.857 & \textbf{\underline{0.899}}     \\
                game              & \textbf{\underline{1.000}}    & 0.901 & 0.895 & 0.999     \\
                gamebiscuit       & \underline{1.000}    & 0.985 & 0.988 & \textbf{\underline{1.000}}     \\
                toycubecar        & \textbf{\underline{0.973}}    & 0.290 & 0.192 & 0.964     \\
                \midrule
                Mean              & \underline{0.987}    & 0.906 & 0.891 & \underline{0.987}     \\
                \bottomrule
            \end{tabular}
            }
            \caption{Fundamental matrices}
            \label{tab:fundamentals}
        \end{subtable}
        \caption{Real datasets AUCs}
    \end{table*}

    
    \section{Conclusion and Future Directions}
        \label{sec:conclusion_and_future_directions}
        We proposed \namealgo, a new algorithm for identifying anomalous samples that do not conform to structured patterns. This was done by wisely coupling -- for the fist time -- two ingredients: (i) embedding input data in the preference space and, (ii) applying isolation-based anomaly detection tool (i.e., \pforest) in the preference space. Our empirical evaluation demonstrated that preference embedding increases the separability between normal (structured) and anomalous (not structured) data, leading to superior performance than simply performing anomaly detection in the original space. Most remarkably, the proposed anomaly-detection method \pforest outperforms all the alternatives where anomaly-detection methods are straightforwardly plugged in the preference space. This result highlights the effectiveness of nested Voronoi tessellations in the isolation process.
We believe there are several research directions to investigate further, including use of non-parametric models for preference embedding (e.g., supervised trained models), and using \namealgo in real-world defect-detection applications. Furthermore, given the ability of \namealgo to deal with arbitrary distance metric, it could be interesting to apply \pforest in spaces other than the preference space.




    \printbibliography
\end{document}



%



